\documentclass[10pt,a4paper]{article}

\usepackage{PRIMEarxiv}

\usepackage[utf8]{inputenc} % allow utf-8 input
\usepackage[T1]{fontenc}    % use 8-bit T1 fonts
\usepackage{hyperref}       % hyperlinks
\usepackage{url}            % simple URL typesetting
\usepackage{booktabs}       % professional-quality tables
\usepackage{amsfonts}       % blackboard math symbols
\usepackage{nicefrac}       % compact symbols for 1/2, etc.
\usepackage{microtype}      % microtypography
\usepackage{lipsum}
\usepackage{fancyhdr}       % header
\usepackage{float}
\usepackage{graphicx}       % graphics
\graphicspath{{media/}}     % organize your images and other figures under media/ folder

\usepackage{fullpage}
\usepackage{graphicx}
\usepackage[utf8]{inputenc}
\usepackage{authblk}

\title{SymBrain: A LARGE-SCALE DATASET OF MRI IMAGES FOR NEONATAL BRAIN SYMMETRY ANALYSIS}

\author[1,2]{Arnaud Gucciardi\thanks{arnaud.gucciardi@toelt.ai}}
\author[2]{Safouane El Ghazouali}
\author[2,3]{Francesca Venturini}
\author[1,4]{Vida Groznik}
\author[2]{Umberto Michelucci}

\affil[1]{University of Ljubljana, Faculty of Computer and Information Science, Ljubljana, Slovenia}
\affil[2]{TOELT llc, Machine Learning Research and Development LAB, D\"ubendorf, Switzerland}
\affil[3]{Institute of Applied Mathematics and Physics, Zurich University of Applied Sciences, Winterthur, Switzerland}
\affil[4]{Faculty of Mathematics, Natural Sciences and Information Technologies, University of Primorska, Koper, Slovenia}

\date{}

\begin{document}

\maketitle

\begin{abstract}
This paper presents an annotated dataset of brain MRI images designed to advance the field of brain symmetry study. 
Magnetic resonance imaging (MRI) has gained interest in analyzing brain symmetry in neonatal infants, and challenges remain due to the vast size differences between fetal and adult brains. Classification methods for brain structural MRI use scales and visual cues to assess hemisphere symmetry, which can help diagnose neonatal patients by comparing hemispheres and anatomical regions of interest in the brain. Using the Developing Human Connectome Project dataset, this work presents a dataset comprising cerebral images extracted as slices across selected portions of interest for clinical evaluation . All the extracted images are annotated with the brain's midline. All the extracted images are annotated with the brain's midline.
From the assumption that a decrease in symmetry is directly related to possible clinical pathologies, the dataset can contribute to a more precise diagnosis because it can be used to train deep learning model application in neonatal cerebral MRI anomaly detection from postnatal infant scans thanks to computer vision. Such models learn to identify and classify anomalies by identifying potential asymmetrical patterns in medical MRI images. 
Furthermore, this dataset can contribute to the research and development of methods using the relative symmetry of the two brain hemispheres for crucial diagnosis and treatment planning.
\end{abstract}

% Keywords

\section{Introduction}

The asymmetry of the brain presents an intriguing paradox. In general, the bodies and brains of most organisms, including humans, exhibit pronounced bilateral symmetry (right and left). This bilateral symmetry is essentially the normative state \cite{bear2020neuroscience}. The human brain, on the other hand, exhibits a distinct bilateral symmetry in shape, but a stunning asymmetry in function \cite{springer_brain_2020, gazzaniga2005forty,kandel2000principles}. The brain's left hemisphere  is predominantly responsible for logical reasoning, language processing, and analytical tasks, while the right hemisphere excels in creative thinking, spatial ability, and holistic processing \cite{bear2020neuroscience}.
In general, brain images obtained from the axial planes of the MRI images show a pronounced bilateral symmetry, a property used in multiple studies to diagnose various brain conditions, such as dementia \cite{herzog2021brain}, to study cognitive processes \cite{rogers_brain_2021}, neural disorders \cite{kalavathi2017review}, and developmental abnormalities or diseases \cite{cara_developmental_2022}.

Magnetic resonance imaging (MRI) is one of the most crucial techniques \cite{dubois2021mri, makropoulos2018review} used to study the brain. In particular, MRI has been used to study brain disorders and development issues in term and preterm infants \cite{prager2007magnetic, dubois2021mri, girard2012mri}. Neonatal cerebral MRI presents several practical and technical challenges. Specifically, the brains of babies and adults differ enormously in terms of size, with the fetal and neonatal brain covering a volume in the range of 100 to 600 ml, in contrast to the average volume of an adult brain of more than 1 liter \cite{allen2002normal}. Additionally, MRI imaging of infants' brains presents unique challenges due to the rapid developmental changes and higher water content, which modify the contrast and clarity of images. Furthermore, the constant motion of infants, including breathing and slight movements, can lead to motion artifacts, complicating image acquisition and analysis. In contrast, adult brain MRI faces challenges primarily related to age-related changes such as brain atrophy and the presence of pathologies, which require a more nuanced interpretation of the images.
For these reasons, despite its importance, the availability of data sources for brain volume MRI for infants is scarce. Among the few available images, there remains a need for comprehensive datasets dedicated to brain symmetry detection, hindering the development and evaluation of automated algorithms for this task. This work addresses this need.

To describe the structures of the brain in newborns and children in development, the classification methods of brain structural magnetic resonance imaging rely on scales and cues that qualitatively assess the symmetry of the brain hemispheres. Scales combine manual subscore assessments on symmetrical regions of the axial view of the brain volume to provide a final evaluation score. The stable reliability of the scale subscores makes it suitable for disease-specific classification questions \cite{fiori2014reliability}, such as cerebral palsy.
Such visual semi-quantitative scales for the classification of brain MRI are applied to children's clinical data as consistent methods to quantify imaging findings in terms of brain lesion severity \cite{arnfield2013relationship}. Inspired by standardized scales with detailed quantitative neuroanatomical characterization to examine the relationship between structure and function in children \cite{fiori2014reliability}, comparison analysis of the hemispheres and anatomical regions of interest in the brain can help diagnose neonatal patients. Such scales have moderately high interrater reliability, supporting their use for further evaluation of automatic symmetry methods and examining the relationship between brain structure and function \cite{fiori2014reliability}. 

Despite the significance of brain symmetry, existing magnetic resonance datasets primarily focus on healthy adults and the general understanding of brain structure. The properties of healthy young adults' brain have been examined and used to describe how the brain typically grows and connects during childhood and the transition through puberty to adolescence and young adulthood. Some popular large-scale projects and datasets, such as the Human Connectome Project (HCP) \cite{hcp} and the Nathan Kline Institute (NKI) \cite{nki}, provide extensive MRI data for studying brain networks and organizational patterns. However, they only concern adults or developing children; they do not specifically target brain symmetry at birth.
The largest available volume dataset for neonatal cerebral images and the one used as the source for this dataset is the Developing Human Connectome Project (dHCP) \cite{hughes2017dedicated}, based on the principles of the HCP, with differences in protocols to adapt to neonatal patients. 

This void in the availability of large-scale brain MRI datasets dedicated to brain symmetry detection motivates the creation of a comprehensive repository of annotated magnetic resonance images designed to facilitate the development and evaluation of automated algorithms for detecting the symmetry axis within brain MRI data. 
%By providing a standardized framework for evaluating brain symmetry, SymBrain offers a significant step forward in advancing understanding of brain structure and functions and improving diagnostic accuracy for various neurological and psychiatric conditions.
The proposed annotated data set presents midline annotations in two-dimensional volume slices. This data set significantly contributes to various applied areas of computer science and clinical areas. The data set is conveniently accessible to train and validate machine learning algorithms to automatically detect midlines in brain MRI images, specifically newborn MRIs. The detection of anomalies or outliers can help radiologists in their diagnosis and save time in image interpretation. Algorithms trained on such datasets can serve as a decision support system, potentially reducing diagnostic errors. , the dataset can be used to study variations in midline structures across different populations, ages, and health conditions, contributing to a better understanding of anatomical variability. It can help medical research, diagnosis, and treatment. Furthermore, the dataset can support research in various medical fields, fostering a deeper understanding of diseases affecting midline structures and facilitating the development of new diagnostic and treatment methods in which researchers can use the dataset to innovate and develop new techniques for image analysis, potentially leading to improvements in automated medical imaging. 
%By making the annotated dataset openly available, it can become a valuable resource for the scientific community, fostering collaboration and accelerating advancements in the field of medical imaging.

\section{Dataset Description}

As previously stated, the dataset created in this paper stems from the dHCP \cite{hughes2017dedicated}, a large-scale effort to map the neural connections in the human brain during development. The project leveraged cutting-edge MRI techniques, including diffusion, structural, and functional MRI, to gather rich datasets from hundreds of participants across different age ranges. The annotations concern the structural MRI available as T1-weighted (T1w) and T2-weighted (T2) MRIs. T1w and T2w are the most commonly used MRI sequence modalities in the clinical visualization field. The T1w and T2w modalities represent two fundamental types of magnetic resonance imaging (MRI) sequences that are pivotal for capturing and visualizing various characteristics of brain tissue and its developmental processes. T1w MRI enhances the signal of fatty tissue, such as brain matter, and suppresses the signal of the water, such as the cerebrospinal fluid located between brain matter and skull. T2w MRI, on the other hand, enhances the signal of the water. 
%T1w images are generated using short echo time (TE) and repetition time (TR) settings. The contrast and brightness of these images predominantly rely on the tissue's T1 properties. Conversely, T2w images are produced by employing longer TE and TR times, where the T2 properties of the tissue primarily determine the contrast and brightness.
The original dHCP data set comprises a combined set of 1050 fetal and neonatal T1w and T2w volumetric scans, precisely 492 T1w and 558 T2w image volumes are accessed. These scans have been completed following a similar protocol in each case and have been collected from diverse populations, including healthy term-born infants, preterm infants, and fetuses with known congenital anomalies. The scans were performed at multiple sites using state-of-the-art MRI machines and protocols specifically designed for neonatal and fetal imaging. All anatomical images for all dHCP subjects have had motion-corrected reconstruction \cite{cordero2016sensitivity}. 
The volume information from the original dataset can be visualized and extracted using different software. This work used the Python package NiBabel \cite{nibabel}. Its API gives full access to header information (metadata) and image volume data as three-dimensional arrays. The raw image volume data from each of the T1w and T2w scans is retrieved as three-dimensional arrays. From the entire array of data, we select the depth of interest for visualization in the coronal view, as these are the views typically used in clinical evaluations \cite{fiori2014reliability}. The volumes are of shape (290, 290, 203) pixels, with the third dimension being the z-axis, the coronal plane along which the cross-sectional slices are selected. 

\section{Materials and methods}

\subsection{Slicing}

\begin{figure}[!ht]
    \begin{center}
        \includegraphics[width=\textwidth]{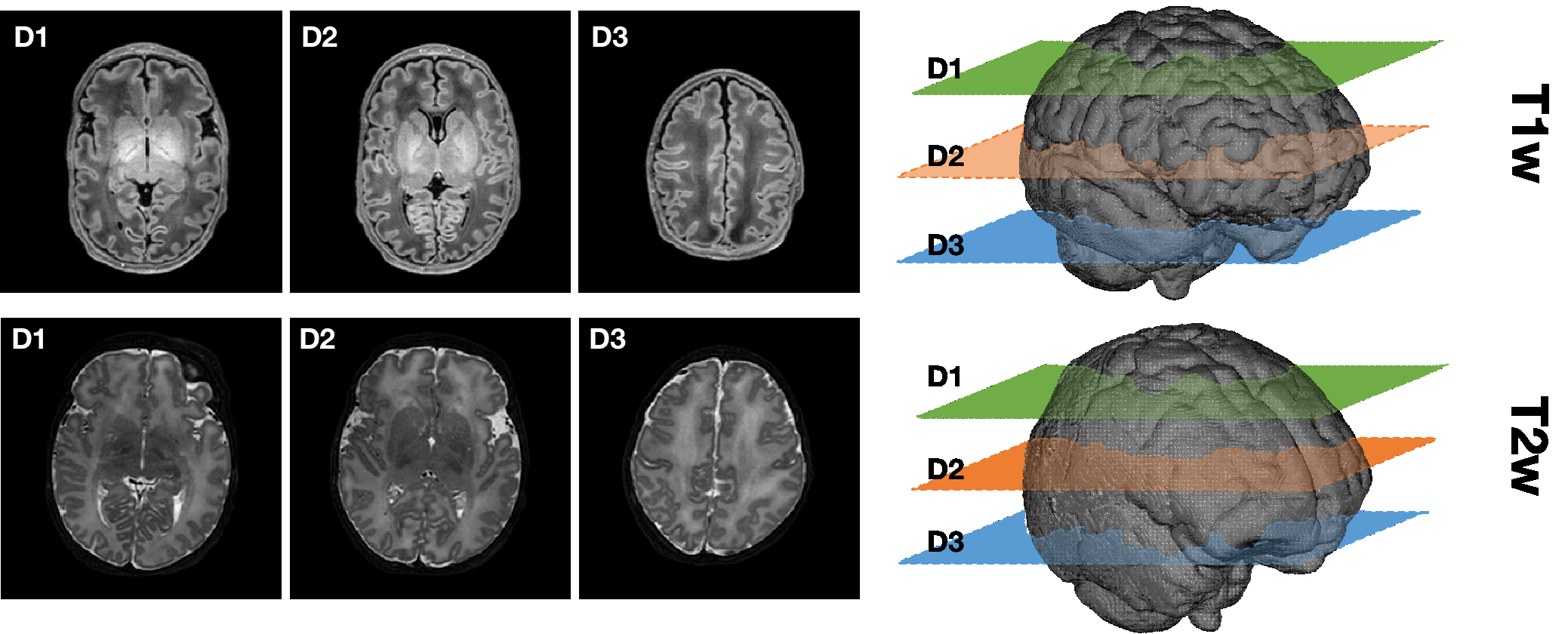}
    \end{center}
    \caption{Representative visualization of the slices on the T1w and T2 volumes, at the three different depths selected. D1: 76px 1, D2: 101px 2, D3: 126px.}
    \label{fig:image_of_the_brain}
\end{figure}

To perform the slicing process on the three-dimensional dHCP volumes, three distinct depths along the vertical (also named frontal or coronal) axis are selected. A two-dimensional cross-section from the three-dimensional image volume is extracted for each depth level, resulting in three two-dimensional cross-sectional slices, per subject (Fig. \ref{fig:image_of_the_brain}).

The three chosen depths on the coronal plane are the D1, D2, and D3 views along the vertical plane, as seen in Fig \ref{fig:image_of_the_brain}. Since the volume shape is (290, 290, 203), the three selected depths represent the axial view at the 101, 76, and 126 z-axis depth (measured in pixels). 
Three two-dimensional sections were extracted from each of the 1050 volumes, creating a total of 3150 two-dimensional axial brain views that can be used for further image analysis. Of the 3150 images, 1476 are T1w-type, and 1674 are T2w-type.
Each slice provides a snapshot of the brain's structure at a particular depth and can be analyzed separately to identify patterns and abnormalities.

The annotation process for medical images can be complex and often time-consuming. The goal is to provide a clear and straightforward labeling process that helps computer vision models learn to identify and classify different features within the images. In the case of brain anomalies, measuring and identifying potential symmetrical patterns in medical MRI images is crucial for diagnosis and treatment planning.

\begin{figure}[H]
    \begin{center}
        \includegraphics[width=\textwidth, scale=0.8]{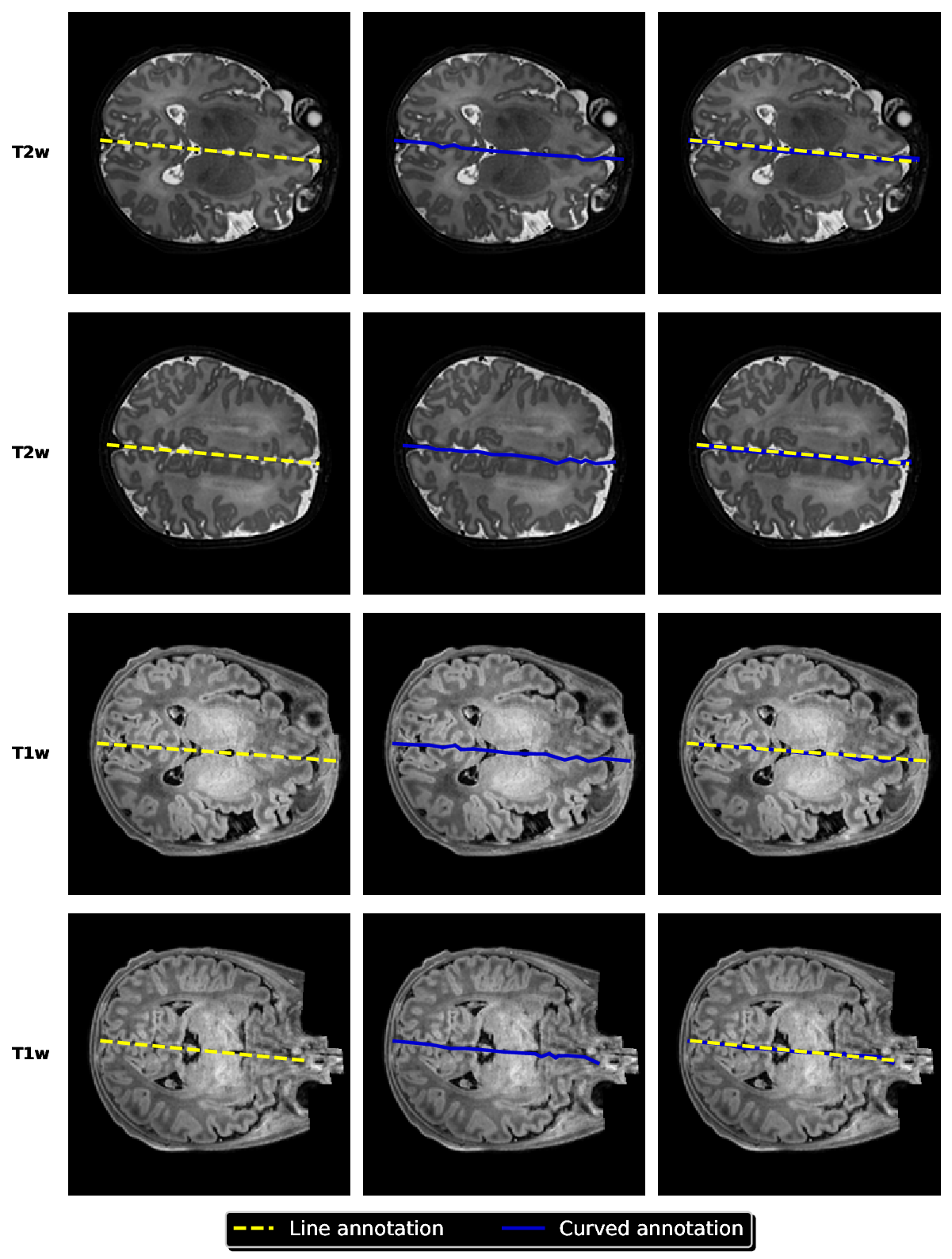}
    \end{center}
    \caption{Comparison of straight and curved midline manual annotation on T1w and T2w slice samples. \textbf{Left}: straight line annotation with two points. \textbf{Center}: curved annotation made of nine control points. \textbf{Right}: visual comparison of the two annotations. Even on a seemingly symmetric brain image, minor curvature changes are visible. }
    \label{fig:image_of_the_annotations}
\end{figure}

To annotate the images, the V7lab annotator software \cite{V7lab} was used to manually draw lines and curves on the images to highlight areas of interest. In the v7lab, the Polyline tool is used to draw a straight line, composed of 2 points, on each slice. 
As illustrated in Fig.\ref{fig:image_of_the_annotations}, these annotations provide a visual representation of the areas of interest within the images. They serve as a rich and detailed reference for the deep learning model, offering precise guidance on the anatomical structures to focus on during the learning process. 

\vspace{15pt} 

\subsection{Annotations}

\textbf{Types of annotations}:

\begin{itemize}
    \item \textbf{Lines of Symmetry}: The first type of annotation involves drawing lines that connect two points on opposite sides of the image, indicating a line of symmetry. This is done by selecting two points on the image, one on each side, and then drawing a straight line connecting them. The coordinates of these two points are recorded along with the corresponding line segment. These line segments serve as a rough approximation of the midline. By comparing the coordinates of the two points, the model can calculate the angle of rotation and other geometric properties of the midline, helping it to better understand the underlying structure.
    \item \textbf{Curved linear paths}: The second type of annotation involves drawing a curved linear path that adapts to the curvature of the midline, providing a more accurate representation of its shape and symmetry. This is achieved by clicking multiple points, with a maximum of ten, along the midline, creating a polyline that closely approximates the true curvature of the interhemispheric fissure. The coordinates of these points are also recorded, allowing the model to analyze the curvature and tortuosity of the midline in greater detail. By examining the sequence of points that form the curved linear path, the model can gain insights into the annotated midline's geometry, such as its radius of curvature, angles of bends, and overall shape.
\end{itemize}

\subsection{Dataset access}
The dataset and annotations are available in the HuggingFace Datasets Hub \cite{hf-dataset}. The Hugginface API allows the loading of the dataset in a single line of code. Additionally, data processing methods are available to quickly get the dataset prepared for training in a deep learning model. The dataset separates the two different modalities into two separate splits. A first split of 1476 rows contains the T1w-type images, and the second split, made of 1674 rows, contains the T2w-type images. Instructions to load the dataset are detailed on the dataset's repository on Hugginface \cite{hf-repo}.

\textbf{Attributes}:
\begin{itemize}
    \item \textit{image}: PIL \cite{clark2015pillow} formatted image representing the cross-section, of shape (290, 290).
    \item \textit{line}: Straight line annotation coordinates on the image, saved as a Python dictionary. ({\textit{x}:x1, \textit{y}:y1}, {\textit{x}:x2, \textit{y}:y2}). Where $(x1,y1)$, $(x2,y2)$ are the starting and end points of the line annotation, in image coordinates.
    \item \textit{radscore}: Radiology score of the volume the image was extracted from. Refer to dHCP documentation \cite{hughes2017dedicated} for scores explanation.
    \item \textit{session}: Session-ID of the original dHCP \cite{hughes2017dedicated} dataset, used for scan identification retrieval.
\end{itemize}

\section*{Data Availability}

The data presented in this study are openly available in Huggingace dataset at https://www.doi.org/10.57967/hf/1372 \cite{hf-repo}, accessed on 12 December 2023.

\section*{Acknowledgments}

Data were provided by the developing Human Connectome Project, KCL-Imperial-Oxford Consortium funded by the European Research Council under the European Union Seventh Framework Programme (FP/2007-2013) / ERC Grant Agreement no. [319456]. We are grateful to the families who generously supported this trial.

\section*{Funding}
This work was supported by the project: ``PARENT'', funded by the European Union's Horizon 2020 Programme MSCA-ITN-2020 Innovative Training Networks Grant Agreement No. 956394.

\section*{Conflict of Interest}
The authors declare no conflicts of interest and no known competing financial interests or personal relationships that could have appeared to influence the work reported in this paper.

\section*{Data Availability}
The SymBrain dataset is openly available in the HuggingFace Datasets Hub \cite{hf-repo}.

\section*{Abbreviations}{
\noindent 
\begin{tabular}{@{}ll}
MRI & Magnetic Resonance Imaging\\
T1w & T1 weighted image \\
T2w & T2 weighted image \\
MSP & Mid-sagittal plane \\
HCP & Human Connectome Project\\
NKI & Nathan Kline Institute\\
dHCP & Developing Human Connectome Project 
\end{tabular}
}

\bibliographystyle{plain}
\bibliography{references}

\begin{thebibliography}{10}

\bibitem{hf-dataset}
Huggingface datasets documentation.
\newblock \url{https://huggingface.co/docs/datasets/index}.
\newblock Accessed: 2023-11.

\bibitem{V7lab}
V7 data engine.
\newblock \url{https://www.v7labs.com/}.
\newblock Accessed: 2023-11.

\bibitem{allen2002normal}
John~S Allen, Hanna Damasio, and Thomas~J Grabowski.
\newblock Normal neuroanatomical variation in the human brain: An mri-volumetric study.
\newblock {\em American Journal of Physical Anthropology: The Official Publication of the American Association of Physical Anthropologists}, 118(4):341--358, 2002.

\bibitem{arnfield2013relationship}
Evyn Arnfield, Andrea Guzzetta, and Roslyn Boyd.
\newblock Relationship between brain structure on magnetic resonance imaging and motor outcomes in children with cerebral palsy: a systematic review.
\newblock {\em Research in Developmental Disabilities}, 34(7):2234--2250, 2013.

\bibitem{bear2020neuroscience}
Mark Bear, Barry Connors, and Michael~A Paradiso.
\newblock {\em Neuroscience: exploring the brain, enhanced edition: exploring the brain}.
\newblock Jones \& Bartlett Learning, 2020.

\bibitem{nibabel}
Matthew Brett, Christopher~J Markiewicz, Michael Hanke, Marc-Alexandre C{\^o}t{\'e}, Ben Cipollini, Paul McCarthy, Dorota Jarecka, CP~Cheng, YO~Halchenko, M~Cottaar, et~al.
\newblock nipy/nibabel: 3.2. 1.
\newblock {\em Zenodo}, 2020.

\bibitem{cara_developmental_2022}
Monica~Laura Cara, Ioana Streata, Ana~Maria Buga, and Dominic~Gabriel Iliescu.
\newblock Developmental {Brain} {Asymmetry}. {The} {Good} and the {Bad} {Sides}.
\newblock {\em Symmetry}, 14(1):128, January 2022.
\newblock Number: 1 Publisher: Multidisciplinary Digital Publishing Institute.

\bibitem{clark2015pillow}
Alex Clark.
\newblock Pillow (pil fork) documentation, 2015.

\bibitem{cordero2016sensitivity}
Lucilio Cordero-Grande, Rui Pedro~AG Teixeira, Emer~J Hughes, Jana Hutter, Anthony~N Price, and Joseph~V Hajnal.
\newblock Sensitivity encoding for aligned multishot magnetic resonance reconstruction.
\newblock {\em IEEE Transactions on Computational Imaging}, 2(3):266--280, 2016.

\bibitem{springer_brain_2020}
Albert Costa.
\newblock {\em The Bilingual Brain: And What It Tells Us about the Science of Language}.
\newblock Springer, 2020.

\bibitem{dubois2021mri}
Jessica Dubois, Marianne Alison, Serena~J Counsell, Lucie Hertz-Pannier, Petra~S H{\"u}ppi, and Manon~JNL Benders.
\newblock Mri of the neonatal brain: a review of methodological challenges and neuroscientific advances.
\newblock {\em Journal of Magnetic Resonance Imaging}, 53(5):1318--1343, 2021.

\bibitem{fiori2014reliability}
Simona Fiori, Giovanni Cioni, Katrjin Klingels, Els Ortibus, Leen Van~Gestel, Stephen Rose, Roslyn~N Boyd, Hilde Feys, and Andrea Guzzetta.
\newblock Reliability of a novel, semi-quantitative scale for classification of structural brain magnetic resonance imaging in children with cerebral palsy.
\newblock {\em Developmental Medicine \& Child Neurology}, 56(9):839--845, 2014.

\bibitem{gazzaniga2005forty}
Michael~S. Gazzaniga.
\newblock Forty-five years of split-brain research and still going strong.
\newblock {\em Nature Reviews Neuroscience}, 6(8):653--659, 2005.

\bibitem{girard2012mri}
Nadine~J Girard, Philippe Dory-Lautrec, M{\'e}riam Koob, and Anca~Melania Dediu.
\newblock Mri assessment of neonatal brain maturation.
\newblock {\em Imaging in Medicine}, 4(6):613, 2012.

\bibitem{hf-repo}
Arnaud Gucciardi.
\newblock mri-sym2 (revision 168e48e), 2023.

\bibitem{herzog2021brain}
Nitsa~J Herzog and George~D Magoulas.
\newblock Brain asymmetry detection and machine learning classification for diagnosis of early dementia.
\newblock {\em Sensors}, 21(3):778, 2021.

\bibitem{hughes2017dedicated}
E.~J. Hughes, T.~Winchman, F.~Padormo, R.~Teixeira, J.~Wurie, M.~Sharma, M.~Fox, J.~Hutter, L.~Cordero-Grande, A.~N. Price, J.~Allsop, J.~Bueno-Conde, N.~Tusor, T.~Arichi, A.~D. Edwards, M.~A. Rutherford, S.~J. Counsell, and J.~V. Hajnal.
\newblock A dedicated neonatal brain imaging system.
\newblock {\em Magnetic Resonance Medicine}, 78(2):794--804, 2017.

\bibitem{kalavathi2017review}
P~Kalavathi, M~Senthamilselvi, and VB~Surya Prasath.
\newblock Review of computational methods on brain symmetric and asymmetric analysis from neuroimaging techniques.
\newblock {\em Technologies}, 5(2):16, 2017.

\bibitem{kandel2000principles}
Eric~R. Kandel, James~H. Schwartz, and Thomas~M. Jessell.
\newblock Principles of neural science.
\newblock {\em McGraw-Hill}, 2000.

\bibitem{makropoulos2018review}
Antonios Makropoulos, Serena~J Counsell, and Daniel Rueckert.
\newblock A review on automatic fetal and neonatal brain mri segmentation.
\newblock {\em NeuroImage}, 170:231--248, 2018.

\bibitem{prager2007magnetic}
Ariel Prager and Sudipta Roychowdhury.
\newblock Magnetic resonance imaging of the neonatal brain.
\newblock {\em The Indian Journal of Pediatrics}, 74:173--184, 2007.

\bibitem{rogers_brain_2021}
Lesley~J. Rogers.
\newblock Brain {Lateralization} and {Cognitive} {Capacity}.
\newblock {\em Animals : an Open Access Journal from MDPI}, 11(7):1996, July 2021.

\bibitem{nki}
Russell~H Tobe, Anna MacKay-Brandt, Ryan Lim, Melissa Kramer, Melissa~M Breland, Lucia Tu, Yiwen Tian, Kristin~Dietz Trautman, Caixia Hu, Raj Sangoi, et~al.
\newblock A longitudinal resource for studying connectome development and its psychiatric associations during childhood.
\newblock {\em Scientific Data}, 9(1):300, 2022.

\bibitem{hcp}
David~C Van~Essen, Stephen~M Smith, Deanna~M Barch, Timothy~EJ Behrens, Essa Yacoub, Kamil Ugurbil, Wu-Minn~HCP Consortium, et~al.
\newblock The wu-minn human connectome project: an overview.
\newblock {\em Neuroimage}, 80:62--79, 2013.

\end{thebibliography}

\end{document}